\documentclass[11pt,a4paper]{article}
\usepackage[hyperref]{emnlp2020}
\usepackage{times}
\usepackage{latexsym}

\usepackage{times}
\usepackage{latexsym}
\usepackage{graphicx} 
\usepackage{url}
\usepackage{multirow}
\usepackage{color}
\usepackage{amsmath}
\usepackage{amssymb}
\usepackage{epstopdf}

\usepackage{enumitem}

\usepackage{microtype}

\aclfinalcopy 
\def\aclpaperid{1702} 

\title{RiSAWOZ: A Large-Scale Multi-Domain Wizard-of-Oz Dataset with Rich Semantic Annotations for Task-Oriented Dialogue Modeling}

\author{Jun Quan\footnotemark[2]  \footnotemark[1] , Shian Zhang\footnotemark[2] \thanks{\ \  Equal Contributions.}, Qian Cao\footnotemark[2], Zizhong Li\footnotemark[3], \and  Deyi Xiong\footnotemark[3] \,\footnotemark[2] \ \\

  \footnotemark[2]\ \  School of Computer Science and Technology, Soochow University, Suzhou, China \\
  \footnotemark[3]\ \  College of Intelligence and Computing, Tianjin University, Tianjin, China \\
  
  \{\tt terryqj0107, sazhang1996, caoqian0905\}@gmail.com,\\ \quad {\tt zizhong1208@126.com}, \quad {\tt dyxiong@tju.edu.cn}
}

\date{}

\begin{document}
\maketitle
\begin{abstract}
In order to alleviate the shortage of multi-domain data and to capture discourse phenomena for task-oriented dialogue modeling, we propose {\bf RiSAWOZ}, a large-scale multi-domain Chinese Wizard-of-Oz dataset with Rich Semantic Annotations. RiSAWOZ contains 11.2K human-to-human (H2H) multi-turn semantically annotated dialogues, with more than 150K utterances spanning over 12 domains, which is larger than all previous annotated H2H conversational datasets. Both single- and multi-domain dialogues are constructed, accounting for 65\% and 35\%, respectively. Each dialogue is labeled with comprehensive dialogue annotations, including dialogue goal in the form of natural language description, domain, dialogue states and acts at both the user and system side. In addition to traditional dialogue annotations, we especially provide linguistic annotations on discourse phenomena, e.g., ellipsis and coreference, in dialogues, which are useful for dialogue coreference and ellipsis resolution tasks. Apart from the fully annotated dataset, we also present a detailed description of the data collection procedure, statistics and analysis of the dataset. A series of benchmark models and results are reported, including natural language understanding (intent detection \& slot filling), dialogue state tracking and dialogue context-to-text generation, as well as coreference and ellipsis resolution, which facilitate the baseline comparison for future research on this corpus.\footnote{The corpus is publicly available at \url{https://github.com/terryqj0107/RiSAWOZ}.}
\end{abstract}

\section{Introduction}

In recent years, we have witnessed that a variety of datasets tailored for task-oriented dialogue have been constructed, such as MultiWOZ \cite{budzianowski}, SGD \cite{rastogi2019towards} and CrossWOZ \cite{zhu2020crosswoz}, along with the increasing interest in conversational AI in both academia and industry \cite{gao2018neural}. These datasets have triggered extensive research in either end-to-end or traditional modular task-oriented dialogue modeling \cite{wenetal2019data,dai2020survey}. Despite of substantial progress made based on these newly built corpora, more efforts in creating challenging datasets in terms of size, multiple domains, semantic annotations, multilinguality, etc., are still in demand \cite{wenetal2019data}.


Among the existing datasets, the majority of them are not large in size, e.g.,  ATIS \cite{hemphilletal1990atis}, WOZ 2.0 \cite{wenetal2017network}, FRAMES \cite{elasrietal2017frames} and KVRET \cite{ericetal2017key}, which might not well support data-hungry neural dialogue models. Very large task-oriented dialogue datasets can be created in a machine-to-machine fashion, such as M2M \cite{shah2018building} and SGD \cite{rastogi2019towards}. Datasets collected in this way need to simulate both user and system and contain unnatural conversations.

\begin{table*}[t]
\centering
\scalebox{0.78}{
\begin{tabular}{l|ccccc|cccc}
\hline
\textbf{Type} & \multicolumn{5}{c|}{\textbf{Single-domain}} & \multicolumn{4}{c}{\textbf{Multi-domain}} \\ \hline
\textbf{Dataset} & \textbf{DSTC2} & \textbf{WOZ 2.0} & \textbf{FRAMES} & \textbf{KVRET} & \textbf{M2M} & \textbf{MultiWOZ} & \textbf{SGD} & \textbf{CrossWOZ} & \textbf{RiSAWOZ (ours)} \\ \hline
Language & EN & EN & EN & EN & EN & EN & EN & ZH & \textbf{ZH} \\
Speakers & H2M & H2H & H2H & H2H & M2M & H2H & M2M & H2H & \textbf{H2H} \\
Domains & 1 & 1 & 1 & 3 & 2 & 7 & 16 & 5 & \textbf{12} \\
Dialogues & 1,612 & 600 & 1,369 & 2,425 & 1,500 & 8,438 & 16,142 & 5,012 & \textbf{10,000} \\
Turns & 23,354 & 4,472 & 19,986 & 12,732 & 14,796 & 115,424 & 329,964 & 84,692 & \textbf{134,580} \\
Avg. turns & 14.5 & 7.5 & 14.6 & 5.3 & 9.9 & 13.7 & 20.4 & 16.9 & \textbf{13.5} \\
Slots & 8 & 4 & 61 & 13 & 14 & 25 & 214 & 72 & \textbf{159} \\
Values & 212 & 99 & 3,871 & 1,363 & 138 & 4,510 & 14,139 & 7,871 & \textbf{4,061} \\
\begin{tabular}[c]{@{}l@{}}Linguistic \\ annotation\end{tabular} & No & No & No & No & No & No & No & No & \textbf{Yes} \\ \hline
\end{tabular}%
}
\caption{Comparison of our dataset to other task-oriented dialogue datasets (training set). H2H, H2M, M2M represent human-to-human, human-to-machine, machine-to-machine respectively.}
\label{conparison table}
\end{table*}


MultiWOZ \cite{budzianowski}, probably the most promising and notable dialogue corpus collected in a Wizard-of-Oz (i.e., Human-to-Human) way recently, is one order of magnitude larger than the aforementioned corpora collected in the same way. However, it contains noisy system-side state annotations and lacks user-side dialogue acts\footnote{In MultiWOZ 2.1, \citet{eric2019multiwoz} re-annotate utterances to fix the noisy state annotation problem via crowdsourced workers.} \cite{eric2019multiwoz, zhu2020crosswoz}. Yet another very recent dataset CrossWOZ \cite{zhu2020crosswoz}, the first large-scale Chinese H2H dataset for task-oriented dialogue, provides semantic annotations on both user and system side although it is relatively smaller than MultiWOZ. The number of domains in both MultiWOZ and CrossWOZ is fewer than 10. MultiWOZ dialogues cover 7 domains. However, the distribution of dialogues over these domains is imbalanced. Dialogues from two domains ({\em hospital}, {\em police}) account for less than 6\% in the training set and don't appear in either the development or test set. CrossWOZ involves 5 domains and dialogue goal descriptions for the domain {\em taxi} and {\em metro} are relatively simple than those from other domains. Neither MultiWOZ nor CrossWOZ provide linguistic annotations to capture discourse phenomena which are ubiquitous in multi-turn dialogues and are important in dialogue modeling \cite{quanetal2019gecor,suetal2019improving,rastogi2019scaling,zhang2019filling}

\begin{figure}[t!] 
\centering 
\includegraphics[scale=0.44]{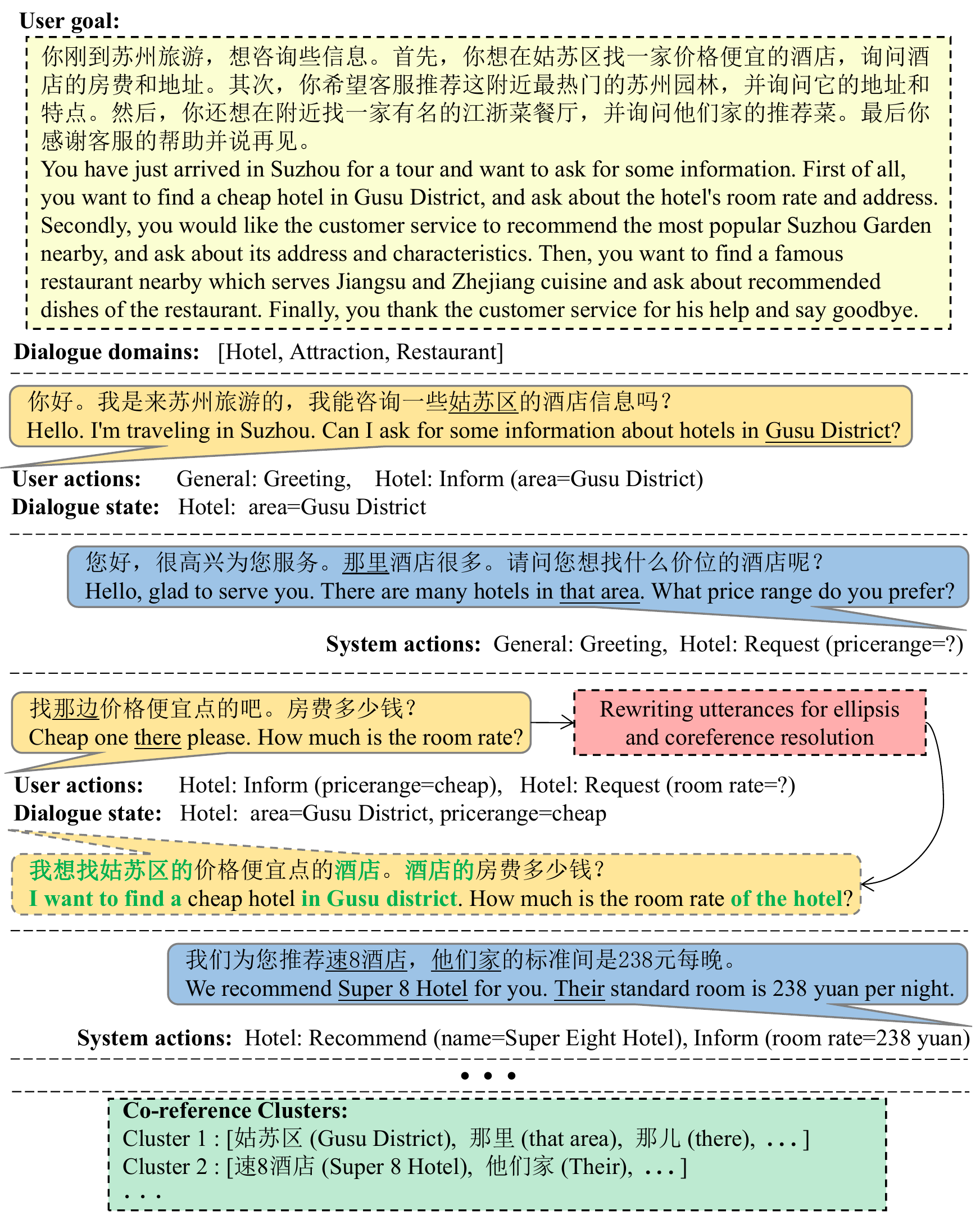}
\caption{A dialogue example spanning over multiple domains. We show dialogue annotations and necessary linguistic annotations (green words for ellipsis resolution and green box for coreference clusters) for each user utterance (in yellow callout) and system utterance (in blue callout). Better viewed in color.}
\label{dialogue_example} 
\end{figure}

In order to alleviate the aforementioned issues, we propose {\bf RiSAWOZ}, a large-scale Chinese multi-domain Wizard-of-Oz task-oriented dialogue dataset with rich semantic annotations. Compared with existing datasets (particularly MultiWOZ and CrossWOZ), our contributions can be summarized as follows:
\begin{itemize}
    \item RiSAWOZ is to date the largest fully annotated human-to-human task-oriented dialogue dataset to our knowledge. It contains 11,200 multi-turn dialogues with more than 150K utterances ranging over 12 domains, namely \emph{Attraction}, \emph{Restaurant}, \emph{Hotel}, \emph{Flight}, \emph{Train}, \emph{Weather}, \emph{Movie}, \emph{TV}, \emph{Computer}, \emph{Car}, \emph{Hospital} and \emph{Education} (particularly after-school remedial courses), several of which are not covered in previous datasets. Compared with other Wizard-of-Oz datasets (e.g. MultiWOZ and CrossWOZ), RiSAWOZ offers a wider and more balanced coverage on both domains and slots, making it suitable for not only practical use in industrial scenarios of related domains but also research on domain adaptation, few/zero-shot learning in task-oriented dialogue. Detailed comparison of RiSAWOZ to existing task-oriented dialogue datasets is shown in Table \ref{conparison table}.  
    
    \item We provide richer manual semantic annotations on the crowd-sourced dialogues, including both dialogue annotations (i.e., various structured semantic labels for dialogue modeling) and linguistic annotations that are not available in previous Wizard-of-Oz datasets (e.g., MultiWOZ or CrossWOZ). Figure \ref{dialogue_example} shows a dialogue example that demonstrates semantic annotations in RiSAWOZ. User goal description, domain label, dialogue states and dialogue acts at both user and system side are annotated for each dialogue. In order to facilitate the study of ellipsis and coreference in dialogue, we also provide two kinds of linguistic annotations collected in two different ways. Annotations for unified ellipsis/coreference resolution via utterance rewriting are more comprehensive and at least one order of magnitude larger than existing datasets with such annotations \cite{quanetal2019gecor,suetal2019improving, zhang2019filling, rastogi2019scaling}. Coreference clusters in each dialogue are also manually annotated, providing a new large-scale coreference dataset on dialogue, which is complementary to previous coreference datasets on documents \cite{pradhanetal2012conll}. In a nutshell, RiSAWOZ integrates human-to-human conversations, dialogue annotations and linguistic annotations on ellipsis/coreference into a single unified dataset.
    
    \item We use RiSAWOZ as a new benchmark testbed and report benchmark results on 5 tasks for future comparison study and tracking progress on this dataset. The 5 tasks are NLU, DST, Dialogue Context-to-Text Generation, Coreference Resolution and Unified Generative Ellipsis and Coreference Resolution. We discuss the usability of the dataset for other tasks, e.g., Dialogue Policy Learning, Natural Language Generation, User Simulator, Dialogue Summarization, etc. The dataset and the benchmark models will be publicly available soon.

\end{itemize}


\section{Related Work}

\label{retated work}
We follow \citet{budzianowski} to roughly categorize existing task-oriented dialogue datasets into three groups: machine-to-machine, human-to-machine, and human-to-human. From the perspective of domain quantity and data scale, most existing datasets cover only one single or a few domains while large-scale multi-domain datasets are not widely available. As suggested by \citet{wenetal2019data}, task-oriented dialogue datasets in languages other than English are few. To the best of our knowledge, there has been no large-scale dialogue datasets with linguistic annotations aiming at ubiquitous discourse phenomena (e.g., ellipsis and coreference) in dialogue. Although some recent works have proposed datasets with utterance completion annotation for ellipsis or coreference in dialogue \cite{quanetal2019gecor, suetal2019improving}, these datasets are at small scale and with simple dialogue goals. No dialogue datasets provide annotations of coreference clusters. 

\textbf{Machine-to-Machine}\quad 
Collecting data of this type requires to create a user and system simulator to generate multi-turn dialogue outlines, which are further transformed into natural language utterances via paraphrasing with predefined rules or crowdsourced workers \cite{shah2018building, rastogi2019towards}. Despite of less human effort required in this approach, the diversity and complexity of created dialogues greatly depend on the quality of user and system simulators. It's also difficult to avoid mismatch between machine-created dialogues and real human conversations.

\textbf{Human-to-Machine}\quad 
In this method, humans converse with an existing dialogue system to collect dialogue data. The Dialogue State Tracking Challenges (DSTC) has provided several datasets created in this way \cite{williamsetal2013dialog, hendersonetal2014second, henderson2014third}. Generally, the quality of human-to-machine data heavily relies on the performance of the given dialogue system.

\begin{table*}[]
\centering
\scalebox{0.75}{
\begin{tabular}{l|cccccccccccc}
\hline
\textbf{Domain} & \textbf{Attraction} & \textbf{Restaurant} & \textbf{Hotel} & \textbf{Flight} & \textbf{Train} & \textbf{Weather} & \textbf{Movie} & \textbf{TV} & \textbf{Computer} & \textbf{Car} & \textbf{Hospital} & \textbf{Courses} \\ \hline
\textbf{Entities} & 40 & 50 & 70 & 665 & 2,016 & 70 & 100 & 105 & 50 & 52 & 17 & 90 \\
\textbf{Slots} & 12 & 11 & 11 & 9 & 10 & 6 & 10 & 11 & 20 & 23 & 16 & 20 \\ \hline
\end{tabular}%
}
\caption{Database statistics on entities and slots.}
\label{Database statistics table}
\end{table*}

\begin{table*}[]
\centering
\scalebox{0.75}{
\begin{tabular}{l|ll}
\hline
\textbf{Domain} & \textbf{Informable Slots} & \textbf{Requestable Slots} \\ \hline
\textbf{Attraction} & name, area, type, consumption, metro station & \begin{tabular}[c]{@{}l@{}}metro station, ticket price, phone number, address,\\ score, opening hours, features\end{tabular} \\ \hline
\textbf{Restaurant} & name, area, cuisine, pricerange, metro station & \begin{tabular}[c]{@{}l@{}}metro station, per capita consumption, address, phone,\\ score, business hours, dishes\end{tabular} \\ \hline
\textbf{Hotel} & name, area, star, pricerange, hotel type, room type, parking & parking, room charge, address, phone, score \\ \hline
\textbf{Flight} & departure, destination, date, class cabin & \begin{tabular}[c]{@{}l@{}}flight information, departure time, arrival time,\\ ticket price, punctuality rate\end{tabular} \\ \hline
\textbf{Train} & departure, destination, date, train type, seat type & \begin{tabular}[c]{@{}l@{}}train number, duration, departure time, arrival time,\\ ticket price\end{tabular} \\ \hline
\textbf{Weather} & city, date & weather condition, temperature, wind, UV intensity \\ \hline
\textbf{Movie} & production country or area, type, decade, movie star & \begin{tabular}[c]{@{}l@{}}director, movie title, name list, release date,\\ film length, Douban score\end{tabular} \\ \hline
\textbf{TV} & production country or area, type, decade, tv star & \begin{tabular}[c]{@{}l@{}}director, TV title, name list, premiere time,\\ episodes, episode length, Douban score\end{tabular} \\ \hline
\textbf{Computer} & \begin{tabular}[c]{@{}l@{}}brand, computer type, usage, memory capacity, screen size,\\ CPU category, pricerange, series\end{tabular} & \begin{tabular}[c]{@{}l@{}}product name, operating system, game performance,\\ CPU model, GPU category, GPU model, features,\\ colour, standby time, hard disk capacity, weight\end{tabular} \\ \hline
\textbf{Car} & \begin{tabular}[c]{@{}l@{}}series, classification, size, number of seats, brand, hybrid,\\ power level, 4WD, fuel consumption, price range\end{tabular} & \begin{tabular}[c]{@{}l@{}}parking assist system, cruise control system,\\ heated seats, ventilated seats\end{tabular} \\ \hline
\textbf{Hospital} & \begin{tabular}[c]{@{}l@{}}name, level, type, public or private, area, general or\\ specialized, key departments, metro station\end{tabular} & \begin{tabular}[c]{@{}l@{}}address, phone, registration time, service time,\\ bus routes, CT, 3.0T MRI, DSA\end{tabular} \\ \hline
\textbf{Class} & grade, subject, level, Day, Time, area & \begin{tabular}[c]{@{}l@{}}campus, start date, end date, start time, end \\ time, times, hours, classroom, teacher, price\end{tabular} \\ \hline
\end{tabular}
}
\caption{All informable and requestable slots in each domain.}
\label{domain slots table}
\end{table*}

\textbf{Human-to-Human}\quad 
To collect data of this type, crowdsourced workers talk to each other according to given dialogue goals to create diverse and natural dialogues. ATIS \cite{hemphilletal1990atis}, WOZ 2.0 \cite{wenetal2017network}, FRAMES \cite{elasrietal2017frames} and KVRET \cite{ericetal2017key} are small-scale datasets built in this way. In contrast, MultiWOZ \citet{budzianowski} and CrossWOZ \cite{zhu2020crosswoz} are two large-scale H2H datasets proposed recently.

\textbf{Coreference Resolution}\quad 
Coreference is ubiquitous in dialogue. However, there is no available dialogue dataset with labeled coreference clusters. Generally, coreference datasets are created on text paragraphs or documents. The OntoNotes 5.0 dataset \cite{pradhanetal2012conll} is one of the most widely-used document-level dataset on coreference resolution from the CoNLL-2012 shared task.

\textbf{Generative Ellipsis and Coreference Resolution}\quad 
In recent years, ellipsis and coreference resolution in dialogue has been treated as an end-to-end generative task. \citet{suetal2019improving} propose to rewrite dialogue utterances to recover all co-referred and omitted information with an annotated Chinese chit-chat dialogue dataset. \citet{quanetal2019gecor} annotate an off-the-shelf task-oriented dialogue dataset CamRest676 \cite{wen2016conditional} with coreference and ellipsis information and propose an end-to-end generative resolution model for both ellipsis and coreference in a single unified framework. This task is also treated as an auxiliary module to improve dialogue understanding \cite{zhang2019filling} and dialogue state tracking \cite{rastogi2019scaling}. However, the scale of these built or used dialogue datasets is relatively small.

\section{Dataset Creation}

\label{dataset creation}
The whole process of data collection consists of database and ontology construction, goal generation, dialogue collection and two rounds of annotations.

\subsection{Database and Ontology Construction}

We crawl 3,325 unique entities with their attributes from several Chinese public websites. Statistics on entities and slots across 12 domains are shown in Table \ref{Database statistics table}. An ontology is constructed over these entities and attributes, which defines all possible slots for each domain and all possible values for each slot. Slots in the dataset can be divided into two categories: informable and requestable, as shown in Table \ref{domain slots table}. Informable slots are attributes that allow the user to constrain the search into the database. Requestable slots represent specific attributes that the user wants to know about an entity.

\subsection{Dialogue Goal}

First of all, we design dialogue goal templates with placeholders representing slots and values. We have designed 80 dialogue goal templates for 12 domains, including 52 single-domain goals and 28 multi-domain goals. Then we randomly sample actual slots and values from the ontology to fill in the placeholders in the templates to generate dialogue goal instances. We finally generate 5,600 dialogue goal instances. An example of dialogue goal (i.e. user goal) is shown at the top of Figure \ref{dialogue_example}. A dialogue goal is a natural language description that only the user can see. The user needs to talk with the system step by step according to the given goal description until the task is finished. We assign each dialogue goal to two different pairs of workers to accomplish. In this way, we can collect two different dialogues from one dialogue goal. The total of dialogues is therefore $2 * 5,600 = 11,200$. This setting can ensure the diversity of collected dialogues while the amount of cost for crowdsourced workers is under the budget.

\subsection{Dialogue Collection and Annotation}

In order to collect high-quality coherent dialogues and annotations, we develop a collecting platform based on the Client-Server architecture, including two versions of client platform for user and system side respectively. Crowdsourced workers can choose to play the role of either user or system. They work in pairs and enter the chat room to construct dialogues. To ensure the quality of dialogue collection, we hire workers via our in-house crowdsourcing platform and train the workers strictly in advance. Finally, we select 92 well-trained workers to participate in our dialogue collection and annotation. At this stage, we collect task-oriented dialogue data with dialogue annotations including domain labels, dialogue states and acts.

\subsubsection{User Side}

During dialogue collection, a user first reads through the natural language description of a given dialogue goal to understand the task that is required to finish. After that, the user communicates with the system step by step to accomplish the given goal. We encourage the users to follow their own personalized language style in communication and train different workers to play the role of user, which makes created dialogues more complex and diverse, and more similar to the spoken conversations in our daily life. According to the instructions described in the given dialogue goal, the user should provide specific constraints to the system step by step and request the corresponding information. The user can terminate the dialogue when he/she believes the task has been accomplished.

\begin{table}[t]
\centering
\scalebox{0.75}{
\begin{tabular}{l|l}
\hline
\textbf{User action type} & Inform, request, greeting, bye, general \\ \hline
\multicolumn{1}{c|}{\textbf{System action type}} & \begin{tabular}[c]{@{}l@{}}Inform, recommend, request, no-offer,\\ greeting, bye, general\end{tabular} \\ \hline
\end{tabular}
}
\caption{Dialogue act types.}
\label{dialogue acts table}
\end{table}

\begin{table*}[htp]
\centering
\scalebox{0.85}{
\begin{tabular}{lcccccc}
\hline
 & Train & Dev & Test & Single-domain & Multi-domain & All \\ \hline
Dialogues & 10,000 & 600 & 600 & 7,261 & 3,939 & 11,200 \\
Turns & 134,580 & 8,116 & 9,286 & 88,618 & 63,364 & 151,982 \\
Tokens & 1,462,727 & 87,886 & 108,032 & 954,298 & 704,347 & 1,658,645 \\
Vocab. & 11,486 & 4,514 & 4,927 & 10,065 & 8,031 & 11,971 \\
Avg. turns & 13.46 & 13.53 & 15.48 & 12.20 & 16.09 & 13.57 \\
Avg. tokens & 10.87 & 10.83 & 11.63 & 10.77 & 11.12 & 10.91 \\ \hline
Avg. acts & 1.46 & 1.46 & 1.54 & 1.43 & 1.51 & 1.46 \\
Avg. u-acts & 1.79 & 1.79 & 1.85 & 1.73 & 1.88 & 1.79 \\
Avg. s-acts & 1.13 & 1.12 & 1.23 & 1.13 & 1.14 & 1.13 \\
Coref. clusters & 17,561 & 1,082 & 1,230 & 10,982 & 10,269 & 20,915 \\
Avg. c-clusters & 1.76 & 1.8 & 2.05 & 1.42 & 2.45 & 1.77 \\ \hline
\end{tabular}
}
\caption{Data statistics. The average numbers of turns and coreference clusters are for each dialogue. The average numbers of tokens and dialogue acts are for each turn. U-acts and s-acts represent user and system acts respectively.}
\label{data statistics table}
\end{table*}

\subsubsection{System Side}

The worker who plays the role of system (i.e. wizard) provides consulting services in various domains to users. When receiving an utterance from the user side, the wizard needs to first determine which domain the user is talking about and convert the user utterance into structured user acts. A dialogue act for both user and system side consists of the act type (i.e. intent) and slot-value pairs such as \emph{inform (area=Gusu District)}. All the dialogue act types are shown in Table \ref{dialogue acts table}. We define 5 different user act types. \textbf{Inform} represents that a user provides specific constraints to information search from the database. \textbf{Request} denotes that a user asks for the values of specific slots. \textbf{Greeting} and \textbf{bye} are to express greeting and farewell. \textbf{General} represents other behaviors that are not covered above. 

By understanding the goal of a user utterance, the wizard needs to annotate the constraints the user wants to provide and the slots requested by the user. The constraints are called belief states which are a set of slot-value pairs. The belief state is persistent across turns and is used to query the database. The wizard then retrieves the database according to the constraints. Considering both the results of database retrieval and the dialogue context, the wizard should send a natural language response to the user. In addition, the wizard needs to convert the natural language system response into structured system acts. 

Similar to the user acts, 7 different system act types are predefined. \textbf{Inform} represents that the system informs the user about the attribute values of specific entities. \textbf{Recommend} denotes that the system recommends required entities to the user. \textbf{Request} represents that the system asks the user whether there are special constraints for the slots in question. \textbf{No offer} means that the system tells the user there is no matched entity. The remaining three act types, \textbf{greeting}, \textbf{bye} and \textbf{general}, are the same as those user act types described above.

Although the tasks for the system side look complex at the first glance, we design a simple and easy-to-operate graphical user interface (GUI). The wizard only needs to follow the prompts to perform simple operations, such as checking the multi-check box, picking the drop-down box, filling in the input box, clicking specific buttons, etc., which can be easily done by well-trained workers.

In this way, the information of domains, belief states, dialogue acts for both user and system side can be annotated during the process of collecting dialogues. A dialogue example with these annotations are demonstrated in Figure \ref{dialogue_example}. 

Different from the data collecting way that multiple workers contribute to one dialogue adopted by MultiWOZ \cite{budzianowski}, in our dataset, the construction and annotation for each dialogue are completed by a pair of well-trained workers. This is to guarantee the coherence and consistency of each created dialogue and the accuracy of the annotation for them. Moreover, we train each worker to play different roles to diversify dialogue utterances.

\subsection{Linguistic Annotation}

\subsubsection{Coreference Clusters Annotation}

We develop a toolkit with easy-to-operate GUI for annotating coreference clusters. With this annotation toolkit, well-trained annotators read through a dialogue to locate all entity mentions. They then group each of these mentions into an appropriate cluster. As shown at the bottom of Figure \ref{dialogue_example}, entity mentions in each cluster are co-referential to one another.

\subsubsection{Ellipsis and Coreference Annotation via Utterance Rewriting}

As shown in Figure \ref{dialogue_example}, both referenced and absent information can be recovered by rewriting an incomplete utterance into a complete version. In this way, we can reformulate ellipsis and coreference resolution as sentence rewriting in a unified framework. The merit of such rewriting is to help the dialogue model better understand the goal of a user utterance in context. 

In order to facilitate such task reformulation, we provide the second type of linguistic annotation on RiSAWOZ: utterance rewriting for ellipsis and coreference resolution. We train crowdsourced workers to accomplish this annotation task and develop an annotation toolkit for them. Each annotator needs to read an entire dialogue sentence by sentence, detecting ellipsis or coreference phenomena in user utterances. For an utterance with ellipsis or coreference, the annotator rewrites the utterance into its complete version with recovered referenced/absent information according to dialogue context. If none of them occurs in the user utterance, the original utterance is kept. Both cases are presented in the example in Figure \ref{dialogue_example}.

\begin{table}[t]
\centering
\scalebox{0.85}{
\begin{tabular}{lcc}
\hline
 & User Utterances & Rate (\%) \\ \hline
Ellipsis & 23,181 & 30.50 \\
Coreference & 19,993 & 26.31 \\
Both & 3,582 & 4.71 \\
Neither & 29,235 & 38.47 \\
Total & 75,991 & 100 \\ \hline
\end{tabular}%
}

\caption{Statistics on utterances containing ellipsis and coreference.}
\label{proportion of ellipsis and co-reference table}
\end{table}

\begin{table*}[t]
\centering
\scalebox{0.83}{
\begin{tabular}{lllccc}
\hline
\textbf{Task} & \textbf{Model} & \textbf{Metrics} & \multicolumn{1}{l}{\textbf{Single-domain}} & \multicolumn{1}{l}{\textbf{Multi-domain}} & \multicolumn{1}{l}{\textbf{All data}} \\ \hline
\multirow{2}{*}{Natural Language Understanding} & BERT & \multirow{2}{*}{Dialogue act F$_1$} & 82.64 & 81.68 & 82.15 \\
 & + context &  & 84.63 & 83.61 & 84.10 \\ \hline
\multirow{3}{*}{Dialogue State Tracking} & TRADE (rand init) & \multirow{3}{*}{Joint Accuracy} & 65.35 & 50.49 &  58.19\\
 & TRADE (fastText) &  & 68.55 & 50.94 &  60.07\\
 & MLCSG (fastText) &  & 73.04 & 58.77 &  66.16\\ \hline
\multirow{4}{*}{Context-to-Text Generation} & \multirow{4}{*}{DAMD} & Inform & 79.19 & 65.00 & 73.73 \\
 &  & Success & 55.66 & 54.68 & 55.18 \\
 &  & BLEU & 32.90 & 22.00 & 27.90 \\
 &  & Combined Score & 100.33 & 81.84 & 92.36 \\ \hline
\multirow{4}{*}{Coreference Resolution} & \multirow{4}{*}{e2e-coref} & MUC F$_1$ & 86.62 & 83.00 & 84.68 \\
 &  & B$^3$ F$_1$ & 83.75 & 79.03 & 81.24 \\
 &  & CEAF$\phi_4$ F$_1$ & 83.11 & 79.78 & 81.29 \\
 &  & Avg. F$_1$ & 84.49 & 80.60 & 82.41 \\ \hline
 \multirow{3}{*}{\begin{tabular}[c]{@{}l@{}}Unified Generative Ellipsis \\ and Coreference Resolution\end{tabular}} & \multirow{3}{*}{GECOR} & EM & 61.65 & 54.61 & 58.26 \\
 &  & BLEU & 87.75 & 86.19 & 87.50 \\
 &  & Resolution\_F$_1$ & 77.00 & 79.15 & 78.14 \\ \hline
\end{tabular}%
}
\caption{Performance of benchmark models on single-domain, multi-domain and all dialogues of test set. }
\label{Performance of Benchmark models}
\end{table*}

\section{Our Dialogue Dataset}

\label{our dialogue dataset}
Our dataset contains not only single-domain dialogues, but also a great amount of multi-domain dialogues where domains are naturally connected. For example, a user wants to travel from one place to another. After checking the air ticket or train ticket, she wants to ask for the local weather information as well. In this section, we will introduce our dataset from two aspects: data structure and data statistics.

\subsection{Data Structure}

As shown in Figure \ref{dialogue_example}, each dialogue in our dataset consists of a user goal description in natural language, a label of dialogue domain, multiple user and system turns and a set of coreference clusters. In each user turn, the user act and dialogue state are annotated over the user utterance. We also label whether there are ellipsis or coreference phenomena in each user utterance. If so, a complete version of the user utterance is provided. In each system turn, the system utterance is labeled with the corresponding system acts.

\subsection{Data Statistics}

\textbf{Dialogue Statistics:} 
We reshuffle all created dialogues and divide them into the training/dev/test sets which maintain approximately the same distribution on domains. As shown in Table \ref{data statistics table}, the training set contains 10,000 dialogues with 134,580 turns. The development and test set contain 600 dialogues with more than 8K and 9K turns respectively. The 5th column of Table \ref{data statistics table} shows the statistics on single-domain dialogues. Multi-domain dialogues (the 6th column of Table \ref{data statistics table}) cover 8 domains excluding \emph{Computer}, \emph{Car}, \emph{Hospital} and \emph{Education}.

After Chinese word segmentation via Jieba,\footnote{https://github.com/fxsjy/jieba} there are 1,658,645 tokens in total in RiSAWOZ. On average, there are 10.91 tokens in each turn and 13.57 turns in each dialogue. Multi-domain dialogues have more turns and utterances are longer than those in single-domain dialogues.

\textbf{Annotation Statistics:}
As shown in Table \ref{data statistics table}, each dialogue contains an average of 1.46 dialogue acts per turn. Each user and system turn have an average of 1.79 and 1.13 dialogue acts respectively. The richness of dialogue acts also make our data set a new challenge. On average, there are 1.77 coreference clusters in each dialogue. As multi-domain dialogues are more complex, each dialogue has an average of 2.45 coreference clusters. Regarding utterance rewriting for ellipsis and coreference resolution, 75,991 user utterances are reformulated, as shown in Table \ref{proportion of ellipsis and co-reference table}. Only 38.47\% of the user utterances have neither ellipsis nor coreference phenomena, and the remaining 61.53\% have at least one of them.






\section{RiSAWOZ as a New Benchmark}

\label{benchmark and analysis}
The large size and rich semantic annotations of RiSAWOZ make it a suitable testbed for various benchmark tasks. We conduct five different evaluation tasks with the benchmark models and in-depth analyses on RiSAWOZ in this section. We also discuss the applicability of RiSAWOZ for other tasks. Results of the 5 tasks are reported in Table \ref{Performance of Benchmark models}.

\subsection{Natural Language Understanding}

\textbf{Task Definition:} 
In task-oriented dialogue system, the NLU module aims to convert the user utterance into the representation that computer can understand, which includes intent and dialogue act (slot \& value) detection.

\noindent \textbf{Model:}
We adapt BERT \cite{devlinetal2019bert} for the NLU task (intent detection and slot filling). We initialize BERT with the Chinese pre-trained BERT model \cite{cui2019pre} and then finetune it on RiSAWOZ. To take dialogue history into account, we employ the same BERT to model previous dialogue context. We also experiment on the situation without context. For fine-tuning BERT on RiSAWOZ, we set the learning rate to 0.00003 and the dropout rate to 0.1.

\noindent \textbf{Results:} 
From Table \ref{Performance of Benchmark models}, we can clearly find that the model using dialogue context preforms better than not. Also, the model obtains lower F$_1$ scores on multi-domain dialogues than single-domain dialogues.

\subsection{Dialogue State Tracking}

\textbf{Task Definition:} Dialogue State Tracking (DST) is a core component in task-oriented dialogue systems, which extracts dialogue states (user goals) embedded in dialogue context. It has progressed toward open-vocabulary or generation-based DST where state-of-the-art models can generate dialogue states from dialogue context directly.

\noindent \textbf{Model:} To report the benchmark results of the DST task, we implement the TRADE model \cite{wuetal2019transferable} and the MLCSG model \cite{quan2020modeling} which improves long context modeling through a multi-task learning framework based on TRADE and achieves the state-of-the-art joint accuracy on the MultiWOZ 2.0 dataset \cite{budzianowski}. 
We train the models with a learning rate of 0.001 and a weight decay rate of 0.5. Early stopping and dropout are also used to prevent overfitting. The dropout rate is set to 0.2.

\noindent \textbf{Results:} 
As illustrated in Table \ref{Performance of Benchmark models}, we show the joint accuracy results for the two models under two different word embedding initialization settings: random and fastText \cite{grave2018learning} initialization. When we use randomly initialized word embeddings of 100 dimensions, TRADE achieves 65.35\%, 50.49\% and 58.19\% joint accuracy on single-domain, multi-domain and all data respectively. While using 300 dimensional pretrained word vectors from fastText, TRADE performs a little better. Under the same setting, MLCSG achieves the higher 73.04\%, 58.77\% and 66.16\% joint accuracy. In general, the performances of the two DST models significantly drop on multi-domain dialogues.

\subsection{Dialogue Context-to-Text Generation}

\textbf{Task Definition:} 
We recast dialogue response generation a sequence-to-sequence problem: encoding dialogue context to decode system response.

\noindent \textbf{Model:}  
To this task, we use the Domain-Aware Multi-Decoder (DAMD) model \cite{zhang2019task} which achieves state-of-the-art performance on the MultiWOZ 2.0 dataset \cite{budzianowski}. It's an end-to-end model proposed to handle the multi-domain response generation problem, which uses one encoder to encode dialogue context and three decoders to decode the belief span, system action and system response. We set the vocabulary size to 8,000 and randomly initialize 50-dimensional word embeddings. The size of hidden states is set to 100. We train the model with a learning rate of 0.005 and a decay rate of 0.5.

\noindent \textbf{Results:} 
As illustrated in Table \ref{Performance of Benchmark models}, we report inform rate, success rate, BLEU \cite{papineni-etal-2002-bleu} and combined score for this task. The inform rate measures the percentage that the output contains the appropriate entity the user asks for, and the success rate estimates the proportion that all the requested attributes have been answered.
The combined score is calculated via $(inform+success) * 0.5 + BLEU$ as an overall quality \cite{zhang2019task}. 
Still, multi-domain dialogues exhibit a high difficulty level.

\subsection{Coreference Resolution}

\textbf{Task Definition:} We predict coreference clusters where all mentions are referring to the same entity for each dialogue.

\noindent \textbf{Model:}  We use the e2e-coref model \cite{lee2017end}, which is the first end-to-end coreference resolution model, as the benchmark model for this task. The model predicts coreference clusters from texts end-to-end without using any auxiliary syntactic parser or hand-engineered mention detector. It considers all spans in a text as potential mentions and learn distributions over all possible antecedents for each mention. The whole process contains two steps: scoring potential entity mentions by calculating embedding representations of corresponding spans and estimating the score for an antecedent from pairs of span representations. The 300 dimensional word vectors from fastText \cite{grave2018learning} are used for the e2e-coref model. We set the size of hidden states to 200 and the number of layers to 3. The model is trained with a learning rate of 0.001 and a decay rate of 0.999. The dropout rate is set to 0.2.

\noindent \textbf{Results:} 
We report the standard MUC, B$^3$ and CEAF$\phi_4$ F$_1$ metrics using the official CoNLL-2012 evaluation scripts and an average F$_1$ score of the three metrics. As shown in Table \ref{Performance of Benchmark models}, the e2e-coref model achieves 84.49\%, 80.60\%, 82.41\% average F$_1$ score on single-domain, multi-domain and all data respectively. The model performs the worst on multi-domain dialogues where coreference links may cross different domains.

\subsection{Unified Generative Ellipsis and Coreference Resolution}

\textbf{Task Definition:} This is a new task reformulated recently \cite{suetal2019improving, quanetal2019gecor}. It usually takes the current user utterance and dialogue context as input. If there exits ellipsis or coreference phenomena in the user utterance, a complete version of the utterance is generated according to the dialogue context. Otherwise, the original user utterance is kept.

\noindent \textbf{Model:} We adopt the GECOR model \cite{quanetal2019gecor} which is an end-to-end generative ellipsis and coreference resolution model with two encoders and one decoder which can produce a pragmatically complete user utterance via generation and copying.
We set both the size of hidden states and word embeddings to 300. We use 300 dimensional fastText \cite{grave2018learning} word vectors to initialize word embeddings in the embedding layer. We train the model with a learning rate of 0.003 and a decay rate of 0.5. Early stopping is used and the dropout rate is 0.5.

\noindent \textbf{Results:} 
We follow \citet{quanetal2019gecor} to use the exact match rate (EM), BLEU \cite{papineni-etal-2002-bleu} and Resolution\_F$_1$ as the evaluation metrics for this task. 
EM measures whether the generated utterances exactly match the ground-truth utterances. Resolution\_F$_1$ is calculated by comparing machine-generated words with ground-truth words only from the ellipsis / co-reference part of user utterances. 
As shown in Table \ref{Performance of Benchmark models}, the GECOR model achieves 58.26\% EM score, 87.50\% BLEU score and 78.14\% Resolution\_F$_1$ score on all data, which are much lower than the results reported on CamRest676 by \citet{quanetal2019gecor}.

\subsection{Other Tasks}

Apart from the five evaluation tasks introduced above, RiSAWOZ can also facilitate the research of many other tasks. For example, the text of dialogues, as well as the annotation of dialogue states and acts, can support the study of dialog policy learning (DPL), natural language generation (NLG) and user simulator. Dialogue act, text and goal description can be potentially used for the task of dialogue summarization \cite{goo2018abstractive}. RiSAWOZ is also suitable for domain adaptation, zero-shot and few-shot learning for multi-domain task-oriented dialogue modeling due to its wide domain coverage. Rich linguistic annotations of RiSAWOZ would also promote the deep integration of discourse and dialogue. We leave these tasks for our future work.

\section{Conclusion}

\label{conclusion}
In this paper, we have presented RiSAWOZ, to date the largest human-to-human multi-domain dataset annotated with rich semantic information for task-oriented dialogue modeling. We manually label each dialogue in RiSAWOZ not only with comprehensive dialogue annotations for various sub-tasks of task-oriented dialogue systems (e.g., NLU, DST, response generation), but also linguistic annotations over ellipsis and coreference in multi-turn dialogue. In addition, the process of data creation and annotation is described in detail. We also report the benchmark models and results of five evaluation tasks on RiSAWOZ, indicating that the dataset is a challenging testbed for future work. RiSAWOZ is featured with large scale, wide domain coverage, rich semantic annotation and functional diversity, which can facilitate the research of task-oriented dialogue modeling from different aspects.

\section*{Acknowledgments}
The present research was supported by the National Key Research and Development Project (Grant No.2019QY1802). We would like to thank the anonymous reviewers for their insightful comments. The corresponding author is Deyi Xiong (dyxiong@tju.edu.cn).

\bibliography{anthology,emnlp2020}
\bibliographystyle{acl_natbib}

\end{document}